\documentclass[letterpaper]{article}
\usepackage{iccc}

\usepackage{times}
\usepackage{helvet}
\usepackage{courier}
\usepackage{graphicx}
\usepackage{url} 
\usepackage{booktabs}
\DeclareGraphicsExtensions{.png,.jpg,.pdf}

\title{IVIE: A Neuro-symbolic Approach to Incremental and Validated Generation of Interactive Fiction Worlds}

\author{Micaela Vaucher $\dagger$ \qquad {\bf Santiago Silveira}\qquad {\bf Santiago Góngora $\dagger$}\qquad {\bf Luis Chiruzzo}\\\\
Instituto de Computación, Facultad de Ingeniería\\
Universidad de la República\\
Uruguay \\ 
$\dagger$ \{micaela.vaucher, sgongora\}@fing.edu.uy
}

\begin{document} 
\maketitle
\begin{abstract}
\begin{quote}
Computational creativity in Interactive fiction faces a fundamental tension: Large Language Models (LLM) may produce creative narratives but struggle with world coherence, while symbolic systems ensure consistency but lack creative flexibility. We present {\em IVIE} (Incremental \& Validated Interactive Experiences), a neuro-symbolic approach to generating complete and playable interactive fiction worlds from scratch. 
Building upon PAYADOR's neuro-symbolic framework, {\em IVIE} implements a four-stage incremental generation pipeline that delegates creative decisions—setting and character creation, puzzle design—to LLMs while grounding the world state through symbolic validation. The system generates worlds with interconnected locations, functional items, non-player characters, and coherent puzzles, all structured around a central goal-oriented architecture. Human evaluation shows the approach generates immersive, thematically coherent worlds with high player engagement. Results seem to indicate that the neuro-symbolic approach successfully balances flexibility with narrative coherence: symbolic validation grounds LLM generation without eliminating generative freedom. However, challenges remain: LLM inconsistencies occasionally bypass puzzle constraints, and objective validation gaps allow some structurally impossible goals. We identify key design considerations for future neurosymbolic interactive storytelling systems, particularly regarding LLM capabilities and their limitations.
\end{quote}
\end{abstract}

\newcommand{\citesubject}[1]{\citeauthor{#1}~(\citeyear{#1})}

\section{Introduction}

Large Language Models can generate fluid narratives on demand, but they constantly fail when asked to maintain a coherent game world across just a few of player turns. Objects vanish and reappear arbitrarily, NPCs forget previous conversations, spatial relationships become contradictory, and puzzle solutions change mid-game~\cite{SkillCheck}. 
Yet the creative potential remains tantalizing: if we could ground LLM generation in symbolic validation, we could significantly advance the generation of consistent interactive narratives.

Interactive fiction (IF), the genre of text-based games in which users explore fictional worlds and influence the narrative through natural-language commands~\cite{RiedlBulitko}, exemplifies this tension between creative flexibility and structural coherence. Classic systems like \emph{Zork}~\cite{Montfort2003} maintained the world state through hand-crafted objects, locations, and deterministic rules, ensuring consistency but requiring extensive manual authoring for each world. 
Procedural content generation (PCG) approaches address this authoring bottleneck by algorithmically creating game content~\cite{Shaker2016}, yet traditional methods for narratives rely on symbolic planning systems that produce rigid, template-based stories lacking creative richness~\cite{Riedl2015}.

LLMs promise a form of computational improvisation, generating fluid narratives on the fly without predetermined scripts. Yet improvisation is not mere spontaneity — it requires technique, memory, and contextual grounding~\cite{Schiaffini2006,Sawyer2000}, precisely what LLMs lack for extended interactions. They hallucinate objects, forget established facts, and lose track of game state~\cite{Castricato2021,Ji2023}. A golden key introduced on turn 5 may be forgotten by turn 25, or worse, the LLM may contradict whether the door was ever locked. For players seeking coherent experiences rather than surreal explorations, such inconsistencies fundamentally break immersion.

Recent neuro-symbolic approaches~\cite{gongora2024payador,Sarker2021} address this by separating concerns: LLMs handle creative generation while symbolic structures maintain the world state.
Systems like PAYADOR~\cite{gongora2024payador} suggest that this works effectively for \emph{playing} in predefined worlds. Extending these architectures to \emph{automatically generate} complete worlds would enable rapid prototyping and procedural variation while maintaining the structural coherence that manual authoring provides.

In this paper, we present \textbf{IVIE} (Incremental \& Validated Interactive Experiences), a neuro-symbolic system that generates complete, playable IF worlds automatically. 
While PAYADOR seemed to indicate that neuro-symbolic architectures can maintain coherence during gameplay, it operated exclusively on manually predefined worlds. 
IVIE extends this framework to world generation itself: rather than hand-crafting each world, IVIE's four-stage incremental pipeline delegates creative decisions—setting and character creation, and puzzle design—to LLMs, while a symbolic layer enforces spatial connectivity, type correctness, and objective solvability at each stage. The source code is available on GitHub\footnote{https://github.com/micaelavaucher/IVIE}.

\section{Related Work}

IF systems face a fundamental challenge: balancing creative narrative flexibility with structural world coherence. Classic parser-based systems like \emph{Zork}~\cite{Montfort2003} solved this through manual authoring: every object, location, and interaction was hand-coded, ensuring consistency but requiring extensive development effort for each world. Modern approaches attempt to escape this authoring bottleneck through different strategies, each with critical limitations.

Systems like AI Dungeon~\cite{Trapova2021} pioneered the idea that LLMs can generate contextually appropriate responses to arbitrary player actions. By concatenating player input with conversation history and generating continuations via LLMs (e.g., GPT-3), these systems enable unprecedented narrative flexibility. However, extensive user testing revealed critical failures~\cite{yong2023playing}: objects vanish without justification and reappear later, NPCs contradict previous statements, game rules become inconsistent as context windows fill, and the model exhibits ``looping" behavior. These issues stem from LLMs' stateless nature—they maintain no persistent world model~\cite{Castricato2021}, relying solely on probabilistic patterns without mechanisms to enforce structural coherence.

Recent work bridges symbolic and neural paradigms in different ways. Some approaches retrieve or maintain narrative context to improve coherence~\cite{chambers2024berall}, or augment game content through data-driven generation~\cite{dagger_2024}, yet these still lack the symbolic grounding necessary to guarantee structural validity. Others delegate creative generation to LLMs while maintaining a structured world state~\cite{Sarker2021}, combining planning with language models~\cite{Kelly2023,zhou2025story2game} or constructing spatial representations dynamically~\cite{Li2025}, but focus on playing or narrating within pre-existing worlds rather than generating complete worlds from scratch.

Our current work IVIE builds directly upon PAYADOR~\cite{gongora2024payador,gongora2026transformations}, a neuro-symbolic framework that separates symbolic world management from neural narrative generation. PAYADOR implements an object-oriented world model (\texttt{Location}, \texttt{Item}, \texttt{Character}, \texttt{Puzzle}, \texttt{Objective}) and employs dual LLMs: a reasoning model interprets player actions to generate structured state transformations, while a narrative model produces scene descriptions~\cite{gongora2025approaches}.
World coherence is maintained through a continuous grounding cycle: when a player inputs an action, the system renders the current symbolic world state as natural language; the reasoning LLM predicts its effects, generating both symbolic transformations (e.g., moving an item, unlocking a passage) and narrative outcomes; these predictions are validated before being applied to update the world state; finally, the narrative LLM generates scene descriptions based on the updated state.
This separation appeared effective at preventing LLM inconsistencies: spatial connectivity is verified through explicit graph structures, object locations are tracked deterministically, and puzzle states persist until explicitly modified. 
However, these systems operate on manually predefined worlds where every location, object, NPC, and puzzle is hand-coded before gameplay.

\section{The IVIE Approach} \label{sec:approach}

IVIE explores the aforementioned limitation through incremental validated generation: a four-stage pipeline that progressively constructs worlds from abstract concepts to concrete playable experiences, with symbolic validation at each stage ensuring spatial connectivity, type correctness, and objective solvability. Given that LLMs can generate narrative text but struggle with structural coherence, while symbolic systems ensure consistency but lack generative flexibility, the central question becomes: \emph{how do we harness both?}

To achieve this separation of concerns, we build upon PAYADOR's neuro-symbolic framework~\cite{gongora2024payador,gongora2026transformations} as our architectural backbone. 
As described previously, PAYADOR maintains world coherence during gameplay through symbolic world state and dual-LLM architecture. 
Building on this framework, IVIE explores the generation of complete worlds automatically, rather than playing within predefined ones.

To achieve this separation of concerns, we build upon PAYADOR's neuro-symbolic framework~\cite{gongora2024payador,gongora2026transformations} as our architectural backbone. Building on this framework, IVIE explores the generation of complete worlds automatically, rather than playing within predefined ones.

IVIE's symbolic component acts as a structured execution and validation layer rather than a creative reasoning engine: it does not generate narrative content, but executes and enforces structural correctness on LLM-proposed state transformations. Rather than asking LLMs to generate creative content and simultaneously maintain world consistency—a task where they demonstrably fail~\cite{gongora2024payador}—we aim to assign each component what it does best:

\begin{itemize}
    \item \textbf{LLMs generate narrative elements:} location descriptions, character backstories, puzzle designs, and objective framing.
    \item \textbf{Symbolic Python structures maintain world state:} following PAYADOR's object-oriented world model; which locations connect to which, what items exist where, which puzzles block which passages, and whether the objective is solvable given the current world configuration.
\end{itemize}

So, the challenge is to generate worlds that are not only creative and thematically coherent, but also structurally valid and where players can complete the objective.

\subsection{Goal-Oriented Architecture}

PAYADOR worlds include objectives that players must complete, like finding items, or reaching locations. We recognized this concept could serve as the organizing principle for IVIE's generation pipeline: rather than generating disconnected elements and hoping they form a coherent world, the objective defines what must exist and how components relate. Every location, item, NPC, and puzzle exists to serve the player's goal, whether finding a hidden artifact, solving a mystery, or escaping a dangerous location. This architecture addresses common PCG challenges by providing the structural constraints necessary to avoid unplayable content~\cite{Bomstrom2016} while maintaining coherent overarching structures across generated elements~\cite{Khaled2013PCG}.

Every generated element exists to serve a central purpose to complete the objective. This objective defines world requirements: If the goal is ``Deliver the ancient scroll to the librarian", the world must contain the scroll as a gettable item, a librarian character positioned in some location, a path connecting the scroll's location to the librarian's location, and potentially puzzles blocking access to either element. Therefore, the generation pipeline works backwards from this objective, determining what elements does this goal require and where should they be placed. Figure~\ref{fig:generated_world_example} shows a generated world opening where all elements are oriented toward the central objective of finding a missing person.

\begin{figure}[t!]
\centering
\fbox{\parbox{0.95\columnwidth}{
\small
\textbf{Generated World Opening}

\textit{Crumbling brick buildings line the street. Fog rolls between the structures, and distant sounds echo through empty alleys. You are a detective investigating mysterious disappearances in the city.}

\textbf{Current Location:} Abandoned Warehouse District

\textbf{Accessible places:} Police Station, Old Docks

\textbf{Blocked passages:} Evidence Room

\textbf{Objective:} Find the Missing Person

\textbf{Items:} Flashlight, Detective's notebook

\textbf{NPC:} Officer Martinez
}}
\caption{Example world opening generated in Generate mode, showing goal-oriented structure where all elements serve the central objective.}
\label{fig:generated_world_example}
\end{figure}

\subsection{Incremental Generation Pipeline}

\begin{figure*}[t]
\centering
\includegraphics[width=\linewidth]{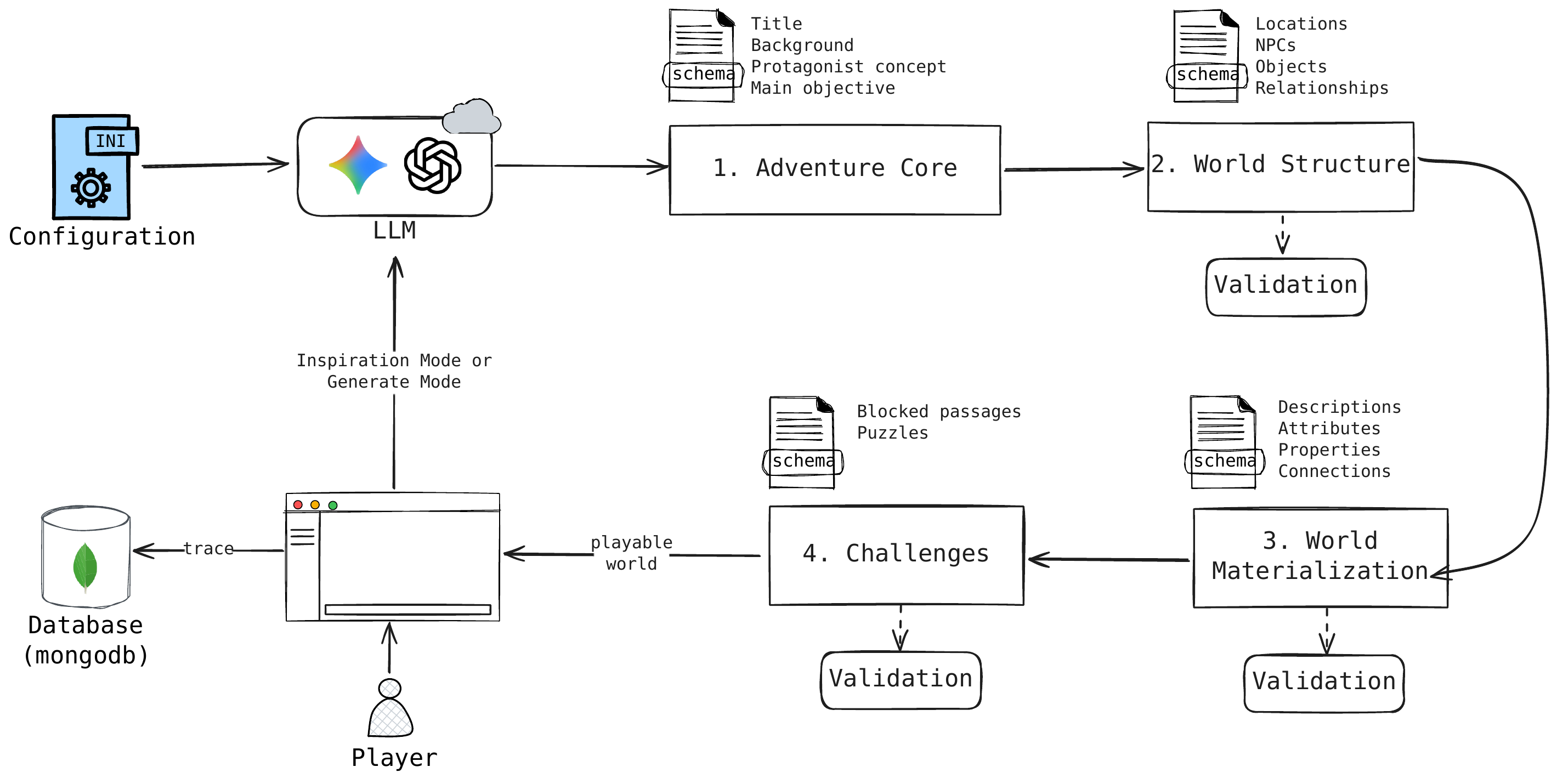}
\caption{IVIE's incremental generation pipeline. Starting from optional user inspiration, the LLM generates worlds progressively across four mandatory stages (Adventure Core → World Structure → World Materialization → Challenges), each with validation gates.}
\label{fig:pipeline}
\end{figure*}

In preliminary experiments, we found that attempting to generate complete worlds in a single LLM call resulted in inconsistencies, as models struggled to maintain coherence across dozens of interdependent elements. Additionally, LLMs would generate worlds of unpredictable sizes, making systematic evaluation and comparison difficult. To address both issues, we designed a four-stage \textbf{incremental generation} pipeline (illustrated in Figure~\ref{fig:pipeline}) with configurable size parameters. IVIE allows designers to specify the number of \texttt{Locations}, \texttt{Items}, \texttt{Characters}, and \texttt{Puzzles} before generation begins, ensuring consistent world scale for controlled evaluation while enabling validation systems to scale appropriately. The pipeline builds worlds progressively across four mandatory stages, from abstract to concrete, with symbolic validation at each boundary. Each stage uses a specialized prompt to guide LLM generation; these prompts are available in the source code repository.

\subsubsection{Stage 1: Adventure Core}

This first stage establishes the narrative foundation by generating three core components: world theme, protagonist description, and objective type. 

IVIE extends PAYADOR's objective system~\cite{gongora2025approaches} with five objective types, each imposing different structural requirements on subsequent stages:

\begin{itemize}
    \item \texttt{REACH\_LOCATION}: Player must navigate to a specific location (e.g., ``Escape the haunted mansion")
    \item \texttt{GET\_ITEM}: Player must obtain a particular item (e.g., ``Retrieve the ancient artifact")
    \item \texttt{DELIVER\_ITEM}: Player must bring an item to a character or location (e.g., ``Deliver the scroll to the librarian")
    \item \texttt{FIND\_CHARACTER}: Player must locate a specific NPC (e.g., ``Find the missing explorer")
    \item \texttt{SOLVE\_MYSTERY}: Player must gather clues to solve a mystery (e.g., ``Discover who stole the crown jewels")
\end{itemize}

These components can be generated in two modes. In \textit{Inspiration mode}, the LLM receives user-provided thematic guidance (e.g., ``mystery in a Victorian mansion") and generates a world aligned with that theme. In \textit{Generate mode}, the LLM fully takes creative responsibility~\cite{colton2012computational}, autonomously inventing theme, setting, and narrative premise without user constraints.\\

\noindent\textbf{Output:} \texttt{WorldConcept} object with adventure \texttt{title}, narrative \texttt{backstory}, \texttt{player\_concept} (character role), and \texttt{main\_objective} (high-level goal description). The formal objective type is determined in subsequent stages.\\
\noindent\textbf{Validation:} This stage does not require validation beyond schema correctness, as the conceptual elements (title, backstory, player concept, objective description) are \texttt{Strings} (purely narrative) and do not impose structural constraints that could render the world unplayable. Any thematically coherent concept is acceptable as input for subsequent stages.

\subsubsection{Stage 2: World Structure}

Once the narrative core is established, this stage identifies the main entities necessary to complete the objective: key \texttt{locations}, relevant \texttt{characters}, and required \texttt{objects}. The LLM generates entity identifiers (names) and brief statements of how each relates to the objective, but without spatial connections, descriptions, or properties. For instance, it might specify ``Ancient Library" and ``Hidden Scroll" as entities, noting the scroll must be found in the library in its description, but not describe what the library looks like, where it connects to, or whether the scroll is locked away. This generated structure works as a world structure to be completed later. In preliminary experiments, we observed that the LLM tended to generate thematically coherent but functionally irrelevant entities (e.g., decorative furniture, background NPCs with no gameplay purpose). To mitigate this, we modified the generation schema to require a \texttt{relevance\_to\_objective} field for each entity, forcing the LLM to justify its functional role explicitly. This ensures every generated element serves the player's path towards objective completion.\\

\noindent\textbf{Output:} \texttt{GeneratedWorld} object containing abstract elements (\texttt{Locations}, \texttt{Characters} and \texttt{Items} without details).\\
\noindent\textbf{Validation:} Schema validation (Pydantic) ensures the output conforms to the expected data structure with correct types and mandatory fields. Size verification confirms if the LLM generated the configured number of elements.

\subsubsection{Stage 3: World Materialization}

This stage completes the world structure from Stage 2 by populating it with the attribute values required by PAYADOR's world model: \texttt{Locations} receive textual descriptions and spatial connections, \texttt{Characters} are assigned descriptions and starting locations, \texttt{Items} are given names and gettability flags, and passages between \texttt{Locations} are established. With entity identifiers already established in Stage 2, the prompt becomes significantly more specific: it explicitly assigns positions to all \texttt{Characters} and \texttt{Items}, and instructs the model not to introduce new entities beyond the existing world structure. These constraints address the previously mentioned problem of thematically coherent but functionally disconnected elements, ensuring every entity has a defined spatial position and all \texttt{Locations} are interconnected.

Mandatory connection rules ensure spatial coherence: (1) if location A connects to B, then B must connect to A (avoiding accidental unidirectional connections that could trap the player), and (2) there must be no isolated locations.
The objective must be completable with the created elements, with at least one logical route from the initial state to its achievement.\\

\noindent\textbf{Output:} Fully instantiated playable \texttt{GeneratedWorld} object.\\
\noindent\textbf{Validation:} Spatial connectivity is verified via depth-first search (DFS), ensuring every \texttt{Location} is reachable from the starting location. Objective completability checks vary by type:

\begin{itemize}
    \item \texttt{GET\_ITEM}: Verifies the target item is both reachable and marked as gettable.
    \item \texttt{REACH\_LOCATION}: Verifies the destination is accessible from the start.
    \item \texttt{DELIVER\_ITEM}: Verifies both the item and the recipient are reachable and the item is obtainable.
    \item \texttt{FIND\_CHARACTER}: Verifies the target NPC exists, and their location is reachable.
    \item \texttt{SOLVE\_MYSTERY}: Verifies all clue items exist in the world and are associated with valid \texttt{Items} that have assigned \texttt{Locations}.
\end{itemize}

\subsubsection{Stage 4: Challenges}

This stage modifies the base world from Stage 3 by adding complexity: puzzles with progressive hint systems and blocked passages between \texttt{Locations}. The LLM receives the complete world structure (all \texttt{Locations}, \texttt{Items}, \texttt{Characters}, and spatial connections) and generates puzzle definitions that specify which passages they block, what solutions unlock them, and hints to guide players toward solutions.

Initially, we attempted to have the LLM generate puzzles where solving one would unlock the next in a specific order, but the model consistently failed with circular references and structural errors, so we simplified the approach. Instead of enforcing a strict sequence in the schema, we prompt the LLM to suggest a ``logical progression" where each challenge narratively builds on previous ones. This appeared to be more effective, as puzzles feel connected without the structural complexity that caused failures.

The generation prompt specifies concrete constraints: each puzzle must connect to either a blocked passage or an essential resource; obstacles and their solutions must be physically separated across different \texttt{Locations}; blocked passages require defined unlock conditions (e.g., obtaining a key, solving a riddle); and solutions must be discoverable through exploration rather than requiring external knowledge.

Each puzzle includes LLM-generated hints organized in a progressive disclosure structure. Hints have 3 levels, starting with general orientation (e.g., ``The answer lies in the study") and progressing to nearly explicit guidance (e.g., ``Examine the portrait of Lady Ashwood"). Additionally, each puzzle includes interaction hints suggesting how to begin (e.g., ``Try talking to the librarian" or ``Examine the locked chest closely"). This hint structure is critical for player assistance, as we discuss in the evaluation section.

Puzzles can be triggered by \texttt{Characters} (e.g., an NPC asks the player to solve a riddle) or by \texttt{Locations} (e.g., examining a room reveals an environmental puzzle). The LLM can strategically redistribute \texttt{Items} to adjust difficulty but cannot create new \texttt{Locations} and must preserve spatial connectivity.\\

\noindent\textbf{Output:} Modified \texttt{GeneratedWorld} object from Stage 3. The world now includes \texttt{Puzzles} with solutions, progressive hints (3-5 levels), and interaction hints; \texttt{BlockedPassages} that lock connections between \texttt{Locations} until puzzle completion; and potentially redistributed \texttt{Items} to create puzzle dependencies.\\
\noindent\textbf{Validation:} Schema validation ensures all puzzle fields are present and correctly typed. Puzzle reward verification checks whether puzzle rewards reference existing \texttt{Items}. If a puzzle grants a non-existent item as reward, the system automatically creates that \texttt{Item} and places it in the puzzle proposer's inventory (for character-proposed puzzles) or in the relevant \texttt{Location} (for environmental puzzles). Finally, spatial connectivity (DFS) and objective completability are revalidated to ensure puzzles and blocked passages do not prevent the player from completing the main objective.

\subsection{Validation Strategy: Correction vs. Rejection}

Each generation stage may fail validation. However, not all validation failures require complete regeneration. IVIE distinguishes between \textit{correctable} issues and \textit{structural failures}. Correctable issues include missing bidirectional connections between \texttt{Locations} or type mismatches in fields—problems that can be automatically fixed without discarding the generated content. Structural failures include unreachable \texttt{Locations} (detected via DFS) or unsolvable objectives (e.g., required \texttt{Items} without assigned positions)—problems that fundamentally break world playability and require regeneration of that stage with enhanced prompts explaining the specific violation.

This distinction aims to balance robustness (rejecting genuinely broken worlds) with efficiency (avoiding unnecessary regeneration when simple corrections suffice). For correctable issues, IVIE applies automatic fixes and proceeds; for structural failures, it provides the LLM with detailed error descriptions to guide regeneration.

Originally, PAYADOR used regex parsing to extract structured data from LLM text responses, which appeared brittle when models generated unexpected formatting. IVIE addresses this by using Pydantic\footnote{\url{https://docs.pydantic.dev/}} models that define strict schemas for each generation stage. Modern LLM APIs enforce these schemas during generation (JSON schema-guided decoding), guaranteeing syntactically valid output. 
Pydantic then performs a second validation pass, checking that generated content is semantically meaningful—catching issues like missing required fields or invalid \texttt{enum} values that pass syntactic checks but would break downstream processing.

For example, the \texttt{GeneratedLocation} model enforces:
\begin{verbatim}
class GeneratedLocation(BaseModel):
    name: str
    connections: List[str]
    blocked_passages: List[BlockedPassage]
    relevance_to_objective: str
\end{verbatim}

If the LLM attempts to omit \texttt{connections} or provide a non-list value, Pydantic raises a validation error. The system attempts regeneration up to three times with enhanced prompts that explicitly describe the validation failure. If all attempts fail, the stage is marked as failed and generation halts, requiring manual intervention or parameter adjustment.

\begin{figure*}[t!]
\centering
\includegraphics[width=0.95\linewidth]{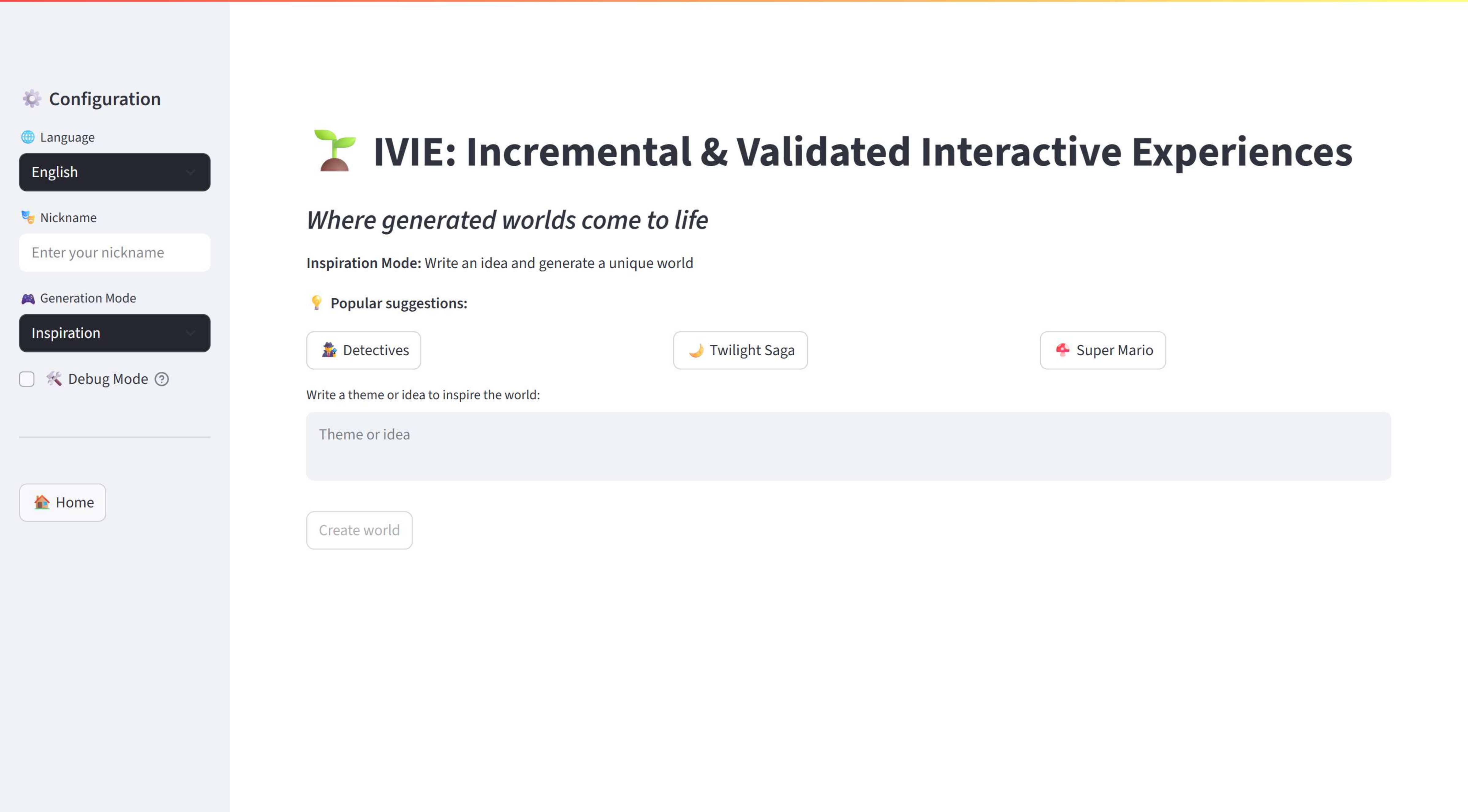}
\caption{IVIE's interface allows users to generate worlds in two modes: Generate (the LLM takes the creative responsibility) or Inspiration (based on thematic input). Configuration options include language, player nickname for identification, and debug mode for inspecting world structure and detecting potential issues.}
\label{fig:ivie_home}
\end{figure*}

\subsection{RAG-Based Memory System}

During gameplay sessions using PAYADOR, each turn receives only the current player action and immediate world state, without access to previous interactions unless explicitly included in the prompt. Moreover, including full conversation history without filtering introduces irrelevant or outdated context that may confuse the model rather than help it. This limitation causes narrative inconsistencies: NPCs may forget previous conversations with the player, or the narration may contradict events from earlier turns.

To address this, IVIE implements Retrieval-Augmented Generation (RAG)~\cite{Lewis2020} with episodic memory. The system stores each gameplay turn as an \texttt{AtomicMemory} unit containing the player action, system-generated narration, and a brief world state summary. For each memory, the system generates 768-dimensional semantic embeddings using Gemini's embedding model\footnote{Google Gemini Embedding API: \url{https://ai.google.dev/gemini-api/docs/embeddings}} and persists them in ChromaDB\footnote{\url{https://www.trychroma.com/}}, which maintains one collection per generated world.

During gameplay, before generating the narration for each new action, the system retrieves the top-3 most semantically similar past memories based on embedding cosine similarity. These retrieved memories are injected into the narrative generation prompt in a dedicated section titled ``Relevant Past Memories," providing the LLM with contextual information about related past events. After the LLM generates narration for the current turn, that turn is stored as a new \texttt{AtomicMemory} and added to the database for future retrieval.

Semantic embeddings enable retrieving conceptually relevant memories even when they share no lexical terms with the current action. For instance, a player input ``examine the portrait" might retrieve a memory where an NPC mentioned the portrait's subject, even though ``portrait" wasn't in that earlier conversation. This can be valuable for narrative coherence, as the LLM can naturally reference past events to create a sense of continuity. Evaluators with RAG enabled reported significantly better long-term narrative coherence, with NPCs remembering previous interactions across dozens of turns, as we discuss in the evaluation.

\subsection{Persistence and Reproducibility}
During development and evaluation, we encountered the critical need to understand how generated worlds behave across complete gameplay sessions. When players reported unexpected behavior, or worlds were not completable, we needed to examine the exact sequence of events: what the player did, how the LLM responded, and how the symbolic world state changed at each step. Additionally, we wanted to enable players to replay interesting worlds from the beginning, exploring different narrative paths within the same generated structure. To address this, IVIE maintains a persistent turn log stored as a JSON array, where each entry captures the full state of that interaction.

Each turn in the array captures: the player's input text, the complete symbolic world state (which \texttt{Locations} are accessible, where \texttt{Items} are located, which puzzles are solved), the LLM's generated narration, and what elements of the world state were modified (e.g., player moved to new \texttt{Location}, item added to inventory, puzzle marked as solved). Turn 0 specially captures the complete initial world state generated by the pipeline, enabling exact reconstruction of the starting conditions. Subsequent turns capture only state changes rather than complete snapshots, reducing storage requirements while maintaining full traceability.

Persisting these generated worlds allow several capabilities: replaying worlds from Turn 0 to explore alternative narrative paths within the same structure, analyzing LLM narrative patterns across complete sessions offline, debugging generation failures by examining the exact state at each turn, and comparing how different players navigate the same generated world. For instance, when evaluators reported incomplete worlds during evaluation, we used the replay capability to load those exact sessions and analyze both the world state and turn history to understand why these failures occurred.

Users interact with all these components through IVIE's web interface (Figure~\ref{fig:ivie_home}), which provides configuration options for generation modes, language selection, and a debug mode that can help visualize the internal world structure and identify potential issues without playing.

\section{Evaluation}

\begin{figure}[t!]
\centering
\includegraphics[width=\columnwidth]{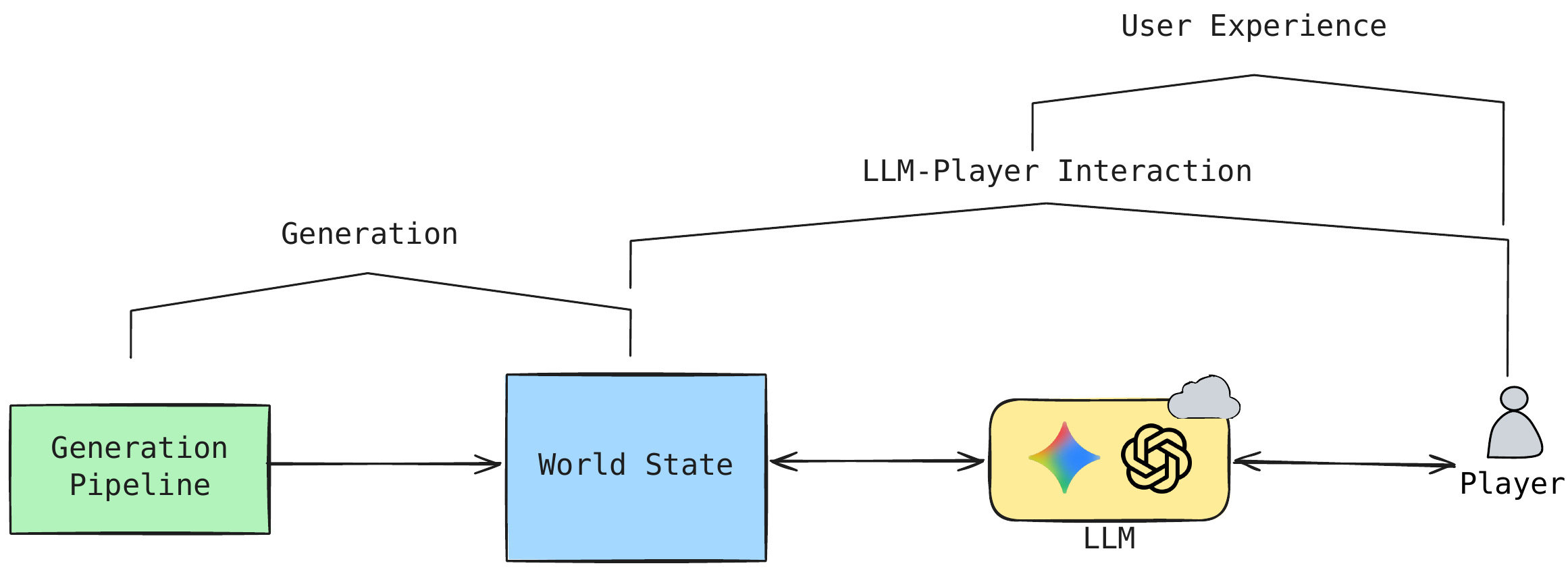}
\caption{Three key dimensions evaluated: World Generation, LLM-Player Interaction, and Player Experience.}
\label{fig:eval_dimensions}
\end{figure}

We evaluated IVIE through human gameplay sessions across three key dimensions (Figure~\ref{fig:eval_dimensions}): \emph{world generation quality} (structural validity, parameter adherence, spatial connectivity), \emph{LLM-player interaction effectiveness} (puzzle difficulty, objective clarity, narrative coherence), and \emph{overall player experience} (engagement, immersion, thematic coherence).

\begin{table}[t!]
\centering
\begin{tabular}{lcc}
\toprule
\textbf{Dimension} & \textbf{Generate} & \textbf{Inspiration} \\
\midrule
Puzzle logic       & 3.88 & 4.00 \\
Hints clarity      & 4.12 & 4.12 \\
Objective clarity  & 4.88 & 4.62 \\
Objective completion & 4.38 & 3.50 \\
NPC purpose        & 3.50 & 3.88 \\
Entertainment      & 4.50 & 4.38 \\
\bottomrule
\end{tabular}
\caption{Mean Likert scores (1--5) per dimension across generation modes (N=8).}
\label{tab:results}
\end{table}

\subsection{Experimental Design}

Eight evaluators with varying gaming experience completed three gameplay sessions. All participants took part voluntarily and evaluation data was anonymized prior to analysis:

\begin{enumerate}
    \item \textbf{Tutorial:} A pre-designed world to familiarize participants with the interface and mechanics.
    \item \textbf{Generate Mode:} A world generated with complete LLM creative freedom.
    \item \textbf{Inspiration Mode:} A world generated based on the participant's favorite IMDB-listed movie.
\end{enumerate}

We distributed participants evenly across experimental conditions to ensure balanced comparison:

\begin{itemize}
    \item \textbf{Language:} 4 Spanish, 4 English
    \item \textbf{Memory System:} 4 with RAG enabled, 4 without RAG
    \item \textbf{LLM Provider:} 4 using Gemini 2.0 Flash, 4 using GPT-4o-mini
\end{itemize}

As previously mentioned, IVIE allows configuration of world-size parameters; in this evaluation, all worlds used identical configurations: 4 \texttt{Locations}, 4 \texttt{Items}, 2 \texttt{Characters} (NPCs), and 1 \texttt{Puzzle}.

After playing, participants completed 5-point Likert scale questionnaires (Yes/Mostly/Sometimes/Almost Never/Never) covering puzzles, objectives, world coherence, and overall experience, plus open-ended qualitative feedback.
The complete evaluation data, including Likert scale responses, full survey questions, and participant gameplay sessions, are available in the source code repository. 
Table~\ref{tab:results} summarizes mean Likert scores per dimension across both generation modes.

Players achieved 100\% objective completion under Generate mode (8/8 worlds) but only 50\% under Inspiration mode (4/8 worlds). 
These Inspiration mode failures revealed critical validation gaps: two worlds passed all validation stages despite having structurally impossible objectives: the required \texttt{Items} existed in the world model but had no assigned \texttt{Location}, making them inaccessible to the player. 
A third failure resulted from API quota exhaustion mid-session\footnote{We used Google Gemini's free tier during evaluation, which has daily request limits.}, which caused the session to terminate abruptly.
Finally, one failure resulted from a persistent LLM reasoning error: despite repeated correct attempts by the player to request an item from an NPC, the reasoning model consistently failed to generate the corresponding transfer transformation, leaving the item in the NPC's inventory and making the objective seem unreachable.

Out of 16 generated worlds, 13 (81.25\%) matched configured parameters exactly, while three worlds (18.75\%) exceeded limits, all in English. Spatial connectivity validation (DFS) succeeded in all cases, with no player becoming isolated in unreachable \texttt{Locations}. Evaluators consistently praised spatial coherence, with one noting that: ``transitions between locations were correct, and narration described surroundings very well.".
Across the 7 evaluated Generate mode worlds, we observed considerable thematic diversity (e.g., fantasy, steampunk, Egyptian mythology, space archaeology, post-plague alchemy) though recurring archetypes emerged (e.g., scholar protagonists, ancient artifact recovery), suggesting that while surface themes vary, LLMs tend to converge on familiar narrative structures. Inspiration mode worlds, by contrast, showed greater narrative diversity, likely driven by the diversity of the source films provided by evaluators.

Evaluators in \emph{Inspiration mode}  consistently reported that the generated worlds strongly reflected the thematic essence of their chosen inspiration movie. We therefore concluded that \emph{Inspiration mode}  achieved 100\% thematic coherence across all evaluated worlds. In contrast, while \emph{Generate mode} received positive feedback on overall creativity, evaluators reported lower perceived utility of generated elements: only 25\% of \emph{Generate mode} evaluators felt all objects and NPCs had a ``real purpose" in achieving the objective, compared to 71\% in \emph{Inspiration mode} . The thematic constraint appears to integrate elements more tightly with objectives, reducing the presence of purely decorative content. We think this hints at a trade-off: \emph{Inspiration mode} produced stronger thematic coherence yet lower completion rates, which might suggest that tighter thematic constraints increase world complexity in ways that expose validation gaps more frequently.

Approximately 75\% of evaluators found the generated puzzles logical with useful hints, but 25\% encountered overly cryptic puzzles requiring specialized knowledge or puzzles with internally inconsistent hints. However, in 3 out of 16 worlds, players were able to bypass puzzles by claiming they had solved them when providing incorrect or no solutions (e.g., typing ``I solve the puzzle'' without actually answering the puzzle). The LLM's reasoning model accepted these claims as valid actions—a phenomenon similar to jailbreaking~\cite{Wei2023Jailbroken}. This issue was previously observed in PAYADOR's original implementation~\cite{gongora2026transformations}, where the reasoning model could suggest state changes (e.g., marking a blocked passage as accessible) that the symbolic engine would accept even when prerequisites weren't met.
This seems to be a fundamental challenge for neuro-symbolic interactive storytelling: validating too strictly may constrain the LLM's generative freedom, while validating too loosely may allow players to bypass puzzle logic entirely.

Regarding the incorporation of RAG for memory, we found that the 4 evaluators with RAG enabled experienced significantly better long-term narrative coherence. For instance, in Evaluator 7's gameplay session, an NPC remembered a successful item theft during turn 17 and referenced it when re-encountered on turn 44 (``Have you come to try and claim what is not yours again?"), suggesting semantic memory retrieval across dozens of intervening turns. However, RAG occasionally mixed contexts and presented puzzles from past interactions instead of the current world state. We believe further exploration of RAG in procedurally generated interactive narratives could address these context-mixing issues through better memory segmentation, for example, organizing memories by type (e.g., location-specific events, NPCs, puzzles) and retrieving only contextually relevant categories based on the current game state. Evaluators with RAG enabled rated entertainment higher on average (4.75 vs. 4.25 without RAG). However, due to the small sample size (N=4 per condition), these results are not included in the table; a more in-depth evaluation would be needed to determine whether RAG has an actual impact on player experience.

Regarding perceived creativity, evaluators responded positively to open-ended questions about world originality in Generate mode, describing worlds as original and engaging to explore. In Inspiration mode, evaluators highlighted the fidelity to source material as a creative achievement in itself, with one noting that the result ``came out of the ordinary and expected.'' We acknowledge, however, that this evaluation does not measure creativity in formal terms nor does it compare against a non-creative baseline. 

Despite technical issues, players found sessions entertaining. Notably, although worlds were small with anticipated 15-25 minute sessions, multiple evaluators played for over an hour, prioritizing narrative exploration over objective completion. Overall, we think this behavior hints at a high level of immersion.

\section{Conclusions and Future Work}

We presented IVIE, a neuro-symbolic system that automatically generates complete and playable IF worlds by strategically delegating creative decisions to LLMs while grounding world coherence through symbolic validation. 
We hope this work represents a step toward addressing the scalability challenges of IF authoring: while manual world design offers artistic control and crafted experiences, automatic generation enables rapid prototyping and increased replayability through procedural variation. IVIE achieves this through a four-stage incremental generation pipeline with symbolic validation at each stage. Unlike purely LLM-based approaches that rely on context window history, IVIE maintains an explicit symbolic world state, enabling validation of structural correctness and prevention of common inconsistencies (such as vanishing objects, contradictory spatial relationships, or unsolvable objectives).

Evaluation results suggest that symbolic validation mechanisms (including DFS for spatial connectivity, Pydantic for type checking, and objective completability verification) can prevent the logical inconsistencies inherent to pure LLM systems while preserving narrative flexibility. Generate mode achieved 100\% completion with high player engagement (sessions exceeding one hour despite small 4-location worlds), and Inspiration mode suggests that 100\% thematic coherence, proving LLMs can generate goal-oriented worlds when properly constrained.

We think IVIE's results offer insights applicable beyond IF to any domain requiring creative generation within structural constraints: the four-stage incremental pipeline with intermediate validation steps suggests that decomposing generation into bounded stages — each with a defined validation contract — may be a viable strategy for grounding open-ended LLM generation in domains where structural correctness matters.

Several directions could significantly improve the approach. Currently, when objective validation fails (e.g., required \texttt{Items} exist but have no assigned \texttt{Location}), the system regenerates the entire stage up to three times before aborting. A more robust approach would automatically correct these issues—detecting unreachable items and assigning them valid positions—similar to how the system currently handles missing puzzle rewards, reducing generation failures and avoiding discarding otherwise playable worlds.

As discussed in the evaluation, the RAG implementation could benefit from organizing memories by type (\texttt{location} events, \texttt{Character} interactions, \texttt{Puzzle} attempts) to address the context-mixing issues we observed. Alternative, specific retrieval techniques for RAG could further improve memory relevance as well.  Additionally, some evaluators encountered overly cryptic puzzles requiring specialized knowledge; implementing an LLM-based difficulty classifier that analyzes puzzle descriptions and hints could identify and regenerate puzzles exceeding reasonable difficulty thresholds.

Also from evaluation feedback, several evaluators expressed interest in deeper NPC interactions. Growing research on LLM role-playing and persona assignment~\cite{tseng-etal-2024-two} suggests that extending \texttt{Character} schemas with explicit personality fields (e.g., cooperative, hostile, enigmatic) could enable richer dialogue and more predictable characterization; currently, NPC behavior depends entirely on LLM generation at each turn without explicit personality guidance. Enabling mid-session save/resume would address a practical need: currently, API failures force players to restart worlds entirely, and sessions cannot span multiple play periods. Implementing true save/resume functionality would require extending the system to initialize world state from arbitrary turn numbers, not just the initial state.

Finally, while IVIE theoretically supports any LLM provider, we only evaluated the approach using commercial APIs (Gemini and GPT-4o-mini). Investigating open-source models (e.g. Llama, Mistral or Qwen) would reduce dependency on proprietary services and enable fully local execution. Exploring specialized smaller models for individual pipeline stages—rather than using one general model for all—could improve both efficiency and generation quality while maintaining local deployability. Beyond generation, IVIE's symbolic world state could enable runtime improvisation: dynamically introducing NPCs, items, or narrative events during gameplay to assist struggling players or increase dramatic tension, similar to how players adapt Tabletop Role-playing Games scenarios in real-time~\cite{martin2016improvisational,SkillCheck}. This would extend IVIE beyond world generation toward real-time narrative improvisation.

\section{Author Contributions}
M.V. and S.S. designed, developed, and conducted the human evaluation of IVIE as part of their Bachelor’s thesis in Computer Science Engineering~\cite{IVIE2025}.
S.G. co-advised that thesis with L.C.; both are also authors of the PAYADOR approach, which IVIE builds upon.

\bibliographystyle{iccc}
\bibliography{iccc}

\end{document}